\documentclass{article}
\usepackage{spconf,amsmath,graphicx}

\usepackage{color, colortbl}

\usepackage{caption}
\usepackage{subcaption}
\usepackage{textpos}

\title{Once-for-All Sequence Compression for Self-Supervised Speech Models}

\name{Hsuan-Jui Chen$^{\dagger }$ \qquad Yen Meng$^{\dagger}$ \qquad Hung-yi Lee \thanks{$^\dagger$Equal contribution.}}
\address{National Taiwan University}

\begin{document}
\maketitle
\begin{abstract}
The sequence length along the time axis is often the dominant factor of the computation in speech processing.
Works have been proposed to reduce the sequence length for lowering the computational cost in self-supervised speech models. However, different downstream tasks have different tolerance of sequence compressing, so a model that produces a fixed compressing rate may not fit all tasks.
In this work, we introduce a once-for-all (OFA) sequence compression framework for self-supervised speech models that supports a continuous range of operating compressing rates. The framework is evaluated on various tasks, showing marginal degradation compared to the fixed compressing rate variants with a smooth performance-efficiency trade-off.
We further explore adaptive compressing rate learning, demonstrating the ability to select task-specific preferred frame periods without needing a grid search.

\end{abstract}

\begin{textblock*}{\textwidth}(0cm, 11.5cm)
\scriptsize\noindent \textbf{Acknowledgement:} We thank the National Center for High-performance Computing (NCHC) of National Applied Research Laboratories (NARLabs) in Taiwan for providing computing and storage resources.
\end{textblock*}

\section{Introduction}
\label{sec:intro}

Large-scale self-supervised speech models \cite{hsu2021hubert,baevski2020wav2vec,chen2022wavlm} have shown their ability to generate state-of-the-art performance across various downstream tasks \cite{yang2021superb}.
Several compression techniques, such as knowledge distillation and weight pruning \cite{hinton2015distilling,frankle2018lottery}, have been studied to lower the ever-growing computational barrier of these large-scale models.
In addition to parameters, the long input of speech accounts for the computation bottleneck as well. To tackle the issue of long sequences, sub-quadratic attention mechanisms have been actively developed to incorporate with Transformers \cite{povey2018time,zaheer2020big}, in order to lower the dependencies of sequence length in memory and runtime. However, these attention mechanisms have various degrees of overhead in practice \cite{tay2020efficient}, which pushes another trend of research: reducing the sequence length itself \cite{lee2022fithubert,wu2022performance,ours,gao2022match}.

Subsampling is a commonly used technique to reduce the sequence length and has been widely adopted in automatic speech recognition (ASR) systems \cite{vanhoucke2013multiframe,miao2015eesen,chan2016listen}.
Shortening sequence within self-supervised speech models has been explored in \cite{lee2022fithubert,wu2022performance} by pairing subsampling with upsampling during the optimization process. Both works demonstrate the effectiveness of computation reduction with subsampling, yet with a limited compressing rate (frame period of 40ms).

Recent works have been proposed to further push the extent of reduction in sequence length for self-supervised speech models \cite{ours,gao2022match}.
Meng et al.\ \cite{ours} proposed variable-length subsampling with guidance from unsupervised phonetic segments, which push the representation duration (average frame period of 90ms) closer to the duration of phone units.
Gao et al.\ \cite{gao2022match} take the data-side approach, truncating audio into shorter segments while retaining the overall duration of audio when pre-training.
Despite the two distinct points of view of these approaches \cite{ours,gao2022match}, both come with similar results showing that shortening the sequence length has different extents of impact on different downstream tasks.
Instead of pre-training a fixed compressing rate model, it would be beneficial if the self-supervised model could support multiple operation points during inference, i.e. assigning task-specific compressing rate.
The idea of on-demand sequence compression has appeared in the work of Vyas et al.\ \cite{vyas2022demand}, where the pre-trained model can select different configurations when evaluating on downstream tasks. However, the approach has only a few discrete operating points with the sequence compressing rate limited to a factor of 2 and is evaluated on a single ASR task.

As the growing research interest in once-for-all (OFA) models \cite{cai2019once,wang2022lighthubert}, we propose a once-for-all sequence compression framework for self-supervised speech models.
To reach a greater compressing rate while having a competitive performance, we build our work on top of Meng et al.\ \cite{ours}, a modified version of DistilHuBERT \cite{chang2022distilhubert} with a variable-length subsampling layer applied.
Our proposed framework allows a sweep-through along a broad spectrum of frame periods while having marginal or no degradation compared to the fixed compressing rate variants.
Evaluation is performed on nine different downstream tasks that are part of the SUPERB challenge \cite{yang2021superb,tsai2022superb}, including content, speaker, and semantics-related tasks.
In addition, with adaptive compressing rate learning, it is possible to obtain an overall best performance without grid search through the whole spectrum.

\begin{figure*}[t!]
    \centering
    \includegraphics[width=0.95\linewidth]{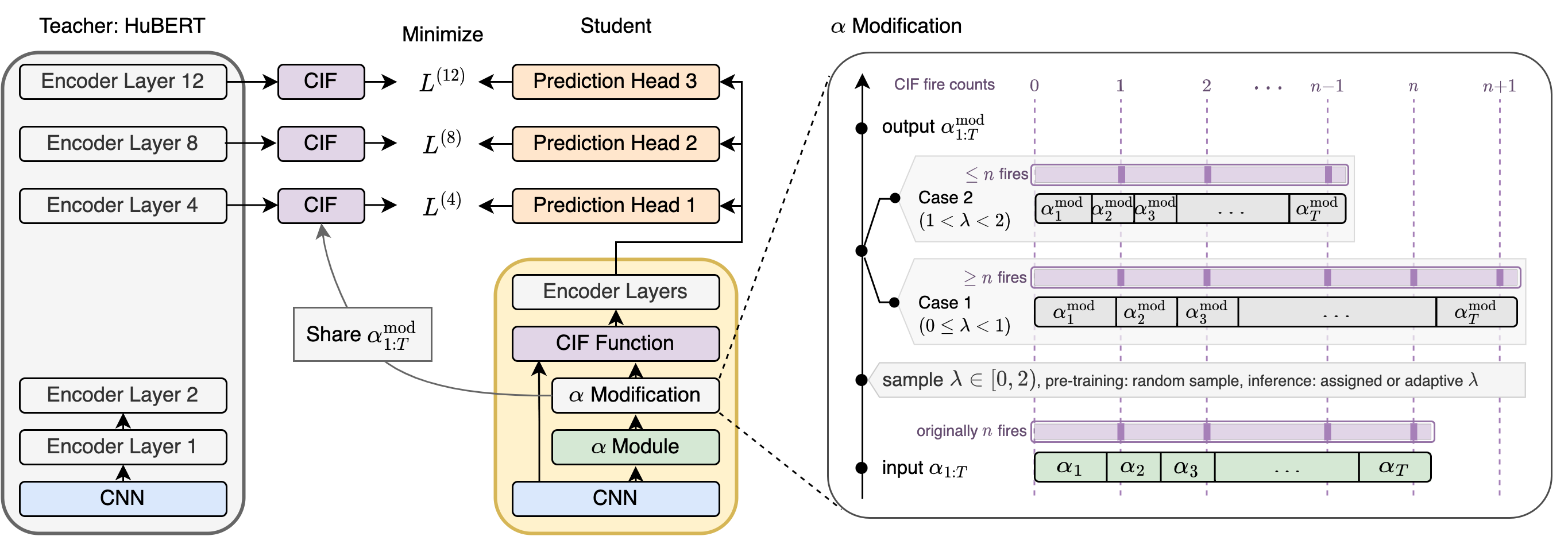}
    \caption{The proposed once-for-all framework. \textit{Left}: Pre-training model architecture. During inference, the prediction heads are discarded. \textit{Right}: Process of $\alpha$ modification. The illustrated length indicates the relative value of $\alpha_i$. By modifying the $\alpha$ weights, the output sequence length will be changed (i.e. number of fires).}
    \label{fig:1}
\end{figure*}

\section{Related Work}

The self-supervised model studied in this paper is DistilHuBERT \cite{chang2022distilhubert}, a distillation framework with a student and a teacher HuBERT \cite{hsu2021hubert} model. Both the student and the teacher models are composed of a CNN feature extractor that converts the input waveform into a sequence of features, followed by a Transformer Encoder.
The student model is trained to minimize the distance between the output representation (through prediction heads) and the chosen layers of the teacher model.

In order to compute the frame-wise reconstruction loss, the representation of the student and the teacher model must have the same number of frames along the time axis, the sequence comes with a frame period of 20ms originally.
In order to further compress the sequence length, Meng et al.\ \cite{ours} proposed to insert a subsampling layer after the CNN feature extractor of the student model. An identical subsample operation is applied to the hidden representation of the teacher model before computing the loss, to match the sequence length of the representations for distillation.

\subsection{Variable-Length Subsampling}
Variable-length subsampling outperforms the fixed-length counterpart (average or convolution pooling) under larger compressing rates.
We follow \cite{ours} to realize variable-length subsampling with Continuous Integrate-and-Fire (CIF) \cite{dong2020cif}.
The CIF module consists of an $\alpha$ module and a CIF function, the $\alpha$ module takes the output sequence from the previous layer and produces a sequence $\alpha_1,\alpha_2,\dots,\alpha_T$ of non-negative numbers, then the CIF function will propose a fire event (output a representation) at time $t$ whenever the accumulated sum up to time $t$, $\sum_{i=1}^t\alpha_i$, crosses an integer boundary (fire threshold).
The value of the output representation at time $t$ will be the weighted sum of the input representation within the current segmentation where the weight values are the corresponding $\alpha$ weights. The length of output representation depends on the number of firing events.
In the case where we have access to phonetic segments, an additional guidance loss can be added to guide the prediction of the CIF module.

\section{Once-for-All Training}

\begin{figure*}[t!]
    \centering
    \includegraphics[width=\linewidth]{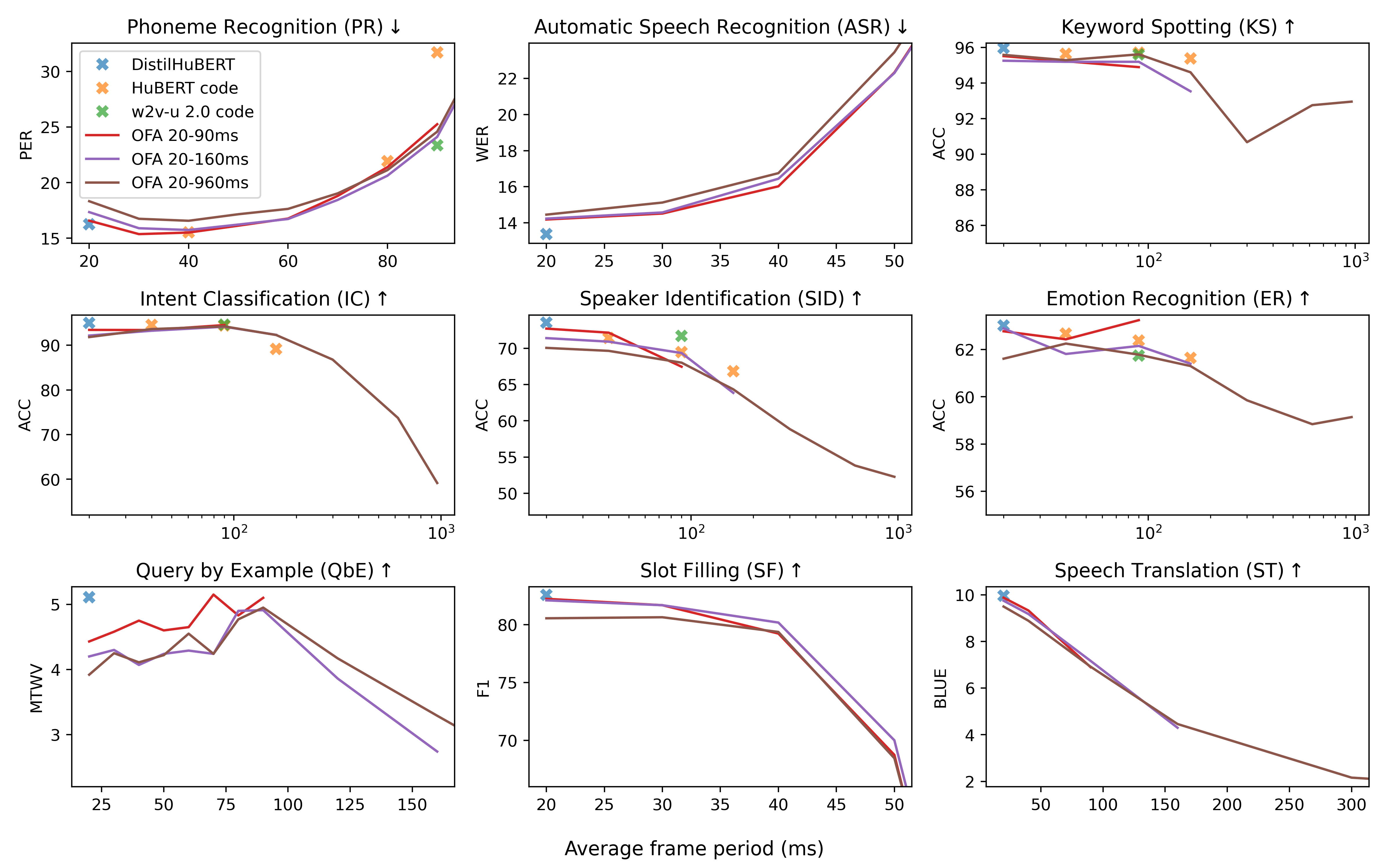}
    \caption{Downstream performance. The horizontal axis shows the average frame period in the unit of milliseconds, the scale is in log scale for KS, IC, SID, and ER.
    Previous works are shown in crosses, including \textit{DistilHuBERT}~\cite{chang2022distilhubert} and variable-length subsampling~\cite{ours}: \textit{HuBERT code} with different smoothing levels and \textit{w2v-u 2.0 code}. The metrics include phone error rate (PER), word error rate (WER), accuracy (ACC), maximum term weighted value (MTWV), F1 score, and BLUE score.}
    \label{fig:2}
\end{figure*}

From the idea of a once-for-all model \cite{cai2019once}, where a model is trained on multiple subtasks or covers a series of subnetworks, to flexibly support various deployment scenarios, we view different compressing rates as different subtasks. Our model is pre-trained on a range of compressing rates to allow on-demand sequence compression in downstream.
The key idea is that we can modify the value of $\alpha_{1:T}$ predicted by the alpha module before feeding into the CIF function.
Below, we discuss how $\alpha$ is modified at the pre-training time and the inference time to realize once-for-all sequence compressing.

\subsection{At Pre-training Time}

A scalar $\lambda$ is introduced, controlling how $\alpha$ will be modified.
At each time step, we randomly sample $\lambda\in[0,2)$. Depending on the value of $\lambda$, modification of $\alpha$ is done as follows.

\vspace{0.5em}
\noindent\textbf{Case 1}\hspace{0.25em}
For $\lambda\in[0,1)$, the modified $\alpha$ is
\begin{align} \label{eq:1}
    \alpha_i^\text{mod}=\lambda\alpha_i+(1-\lambda).
\end{align}

\vspace{0.5em}
\noindent\textbf{Case 2}\hspace{0.25em}
For $\lambda\in[1,2)$, the modified $\alpha$ is
\begin{align} \label{eq:2}
    \alpha_i^\text{mod}=\frac{(2-\lambda)\alpha_i}{\min\left((2-\lambda)\sum_{i=1}^T\alpha_i, 1\right)},
\end{align}
the numerator term linear scales the $\alpha$ from $1$ to $0$, and the denominator term ensures that the CIF function locates at least one boundary for an utterance.
At $\lambda=1$, both Equation \ref{eq:1} and \ref{eq:2} are reduced to $\alpha_{1:T}^\text{mod}=\alpha_{1:T}$,
thus a piecewise linear function $F$ can combine both equation as $\alpha_{1:T}^\text{mod}=F(\alpha_{1:T},\lambda ; \lambda\in[0,2))$, while ensuring the differentiability.

The $\alpha$ modification is essentially up-scaling or down-scaling of the original predicted $\alpha$ weights. 
Since the original value of $\alpha_{1:T}$ is mapped to 0 to 1 via a Sigmoid function, we would like the modified $\alpha_{1:T}^\text{mod}$ to stay in the same range to prevent multiple firing events within a single time step, which will result in duplicated output frames.
In the case of up-scaling, naive scaling will result in some time steps having $\alpha$ weights larger than 1, inevitably producing duplicated frames. Thus for up-scaling, the modification is done by interpolating $\alpha$ weights with ones, as in Equation \ref{eq:1}.
However, Equation \ref{eq:1} can not be extended to down-scaling, which will result in negative $\alpha$ weights, hence, naive scaling is used when down-scaling, as in Equation \ref{eq:2}.
As a result, to achieve a larger compressing range, two cases are required.

To summarize, in the case of $\lambda=0$, the model is equivalent to the vanilla DistilHuBERT, where each timestep fires a representation. In the case of $\lambda=1$, the model is reduced to a single model in Meng et al.
In the case that $\lambda$ is extremely close to 2, i.e. $\lambda\rightarrow 2^-$, the model outputs a single representation per input utterance.
The entire once-for-all pre-training framework and the $\alpha$ modification process are illustrated in Figure~\ref{fig:1}.
The guidance loss proposed in Meng et al.\ is necessary for this framework and is applied to the original $\alpha_{1:T}$.

\begin{figure*}[t!]
    \centering
    \includegraphics[width=\linewidth]{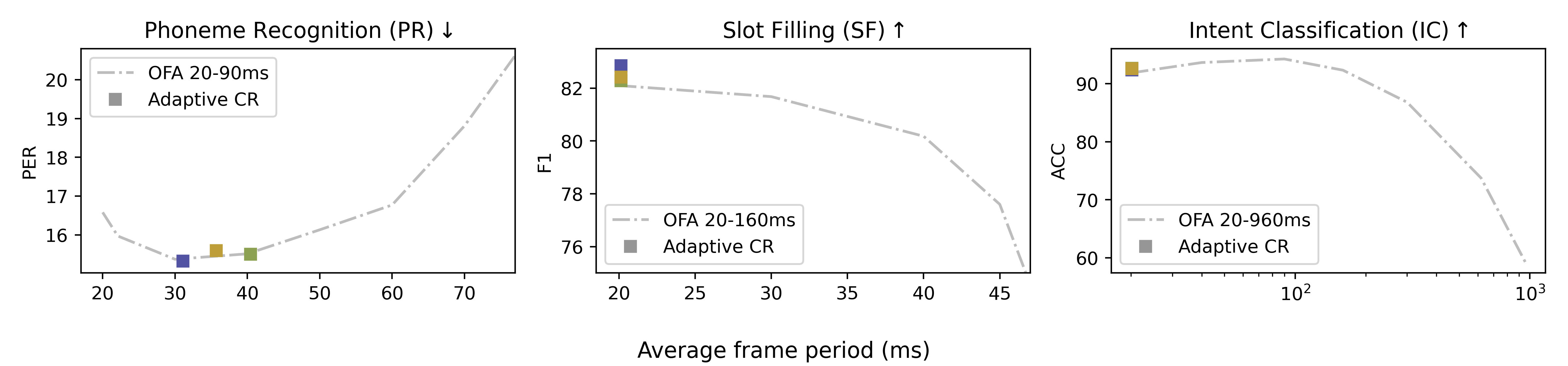}
    \caption{Results of adaptive compressing rate (CR) learning. Different colors denote different initial compressing rates. The initial compressing rates are randomly sampled from the compressing range of each pre-trained model.}

    \label{fig:3}
\end{figure*}

\subsection{At Inference Time}
At inference time, one can choose any value of $\lambda\in[0,2)$. 
Each distinct value of $\lambda$ is associated with a specific compressing rate.
The relation between $\lambda$ and the compressing rate depends on the pre-extracted segments used for guidance loss.
Since the variable $\lambda$ is a continuous and differentiable variable that controls the overall sequence length, one can also treat it as a downstream parameter, and adaptively learn the sequence length with downstream tasks. Additional soft constraints on $\lambda$ can be added to control the compressing rate of the model.

\section{Experiments}

Our experiment is based on the DistilHuBERT implementation with the S3PRL toolkit \cite{yang2021superb}.
The pre-training hyperparameters, including the learning rate schedule, batch size, and distilled layers, are the same as the original implementation.
The model is pre-trained on the 960-hour LibriSpeech \cite{panayotov2015librispeech} dataset.
We follow \cite{ours} for the setting of the $\alpha$ module used in CIF, and generate the segments used for guidance loss from the unsupervised ASR \cite{liu2022towards} trained on 100-hour LibriSpeech and texts from the LibriSpeech language modeling text.

Our framework is evaluated on a subset of the SUPERB benchmark, including phoneme recognition (PR), automatic speech recognition (ASR), keyword spotting (KS), intent classification (IC), speaker identification (SID), emotion recognition (ER), query by example spoken term detection (QbE), slot filling (SF), and speech translation (ST). Only the last layer is used for downstream evaluation.

\subsection{Once-for-All Sequence Compression}
Three pre-trained models with different sampling ranges are tested.
The models uniform sampled $\lambda$ from $\lambda\in[0,1]$, $[0,1.5]$, and $[0,2)$ at pre-training time, and with the segment guidance from the unsupervised ASR, they have a compressing range of 20-90ms, 20-160ms, and 20-960ms respectively.
The downstream results are shown in Figure~\ref{fig:2}.

Since our models come with a continuous range of compressing rates, we are able to finely sweep through frame periods up to 960ms.
We observe that it is possible to further push the compressing rate for utterance-level tasks, namely KS, IC, SID, and ER, e.g. the KS task retains an accuracy of 92\% with a large frame period of nearly 1 second.
In addition, the once-for-all model can be evaluated on all the downstream tasks, including tasks that require a smaller frame period, such as ASR and SF, which fails to be performed by a single model with a fixed and larger compressing rate.

In general, the model pre-trained on a smaller compressing range performs slightly better overall. However, the model pre-trained on the largest compressing range still has a comparable performance with the models pre-trained on a fixed compressing rate from previous works while having the versatility of covering a larger spectrum of compressing rates.
The efficiency improvement is consistent with the reduction in sequence length, and little overhead is introduced by the CIF module. Specifically, the multiply-accumulate operations (MACs) reduction in the Transformer layers including the CIF module is 68.7\% and 90.6\% with frame periods around 90ms and 960ms, respectively.

\subsection{Adaptive Compressing Rate Learning}

As we can select different compressing rates from a continuous range, grid search is a great way to have an overview of the whole landscape. However, instead of grid search through frame periods, it is also possible to obtain a compressing rate that performs close to the overall best result with less effort.

To realize adaptive compressing rate learning, we would like the compressing rate to be learnable while staying in the same range as pre-training. Thus, an additional parameter, serving as the trainable $\lambda$, is introduced and mapped to the same sample range as pre-training via a Sigmoid function.
Since each task has different characteristics (converge speed etc.), adaptively learning the compressing rate requires some task-specific tuning.
The trainable $\lambda$ is optimized with an SGD optimizer with the momentum set to 0.9 and the learning rate set to 1e-3, 1e-2, and 1e-2 for PR, SF, and IC, respectively.
The results are shown in Figure~\ref{fig:3}. Despite the compressing rate adaptively learned does not generate the overall best result in every case, the outcome is on par with the best result using grid search.

\section{Conclusion}
In this work, we propose a once-for-all sequence compression framework for self-supervised speech models.
The proposed once-for-all pre-trained model produces comparable performance with each fixed compressing rate model from previous works under the same frame periods while having the advantage of selecting different compressing rates with the best trade-off during inference considering either the computing resource limitation or task-specific requirements.
In addition, with adaptive compressing rate learning, an overall best result for a specific downstream can be obtained without the effort of grid search.

\bibliographystyle{IEEEbib}
\bibliography{strings,refs}

\begin{thebibliography}{10}

\bibitem{hsu2021hubert}
Wei-Ning Hsu, Benjamin Bolte, Yao-Hung~Hubert Tsai, Kushal Lakhotia, et~al.,
\newblock ``{HuBERT}: Self-supervised speech representation learning by masked
  prediction of hidden units,''
\newblock {\em IEEE/ACM Transactions on Audio, Speech, and Language
  Processing}, vol. 29, pp. 3451--3460, 2021.

\bibitem{baevski2020wav2vec}
Alexei Baevski, Yuhao Zhou, Abdelrahman Mohamed, and Michael Auli,
\newblock ``wav2vec 2.0: A framework for self-supervised learning of speech
  representations,''
\newblock {\em Advances in Neural Information Processing Systems}, vol. 33, pp.
  12449--12460, 2020.

\bibitem{chen2022wavlm}
Sanyuan Chen, Chengyi Wang, Zhengyang Chen, Yu~Wu, et~al.,
\newblock ``Wavlm: Large-scale self-supervised pre-training for full stack
  speech processing,''
\newblock {\em IEEE Journal of Selected Topics in Signal Processing}, 2022.

\bibitem{yang2021superb}
Shu-wen Yang, Po-Han Chi, Yung-Sung Chuang, Cheng-I~Jeff Lai, et~al.,
\newblock ``{SUPERB}: Speech processing universal performance benchmark,''
\newblock in {\em Interspeech}, 2021, pp. 1194--1198.

\bibitem{hinton2015distilling}
Geoffrey Hinton, Oriol Vinyals, and Jeffrey Dean,
\newblock ``Distilling the knowledge in a neural network,''
\newblock in {\em NIPS Deep Learning and Representation Learning Workshop},
  2015.

\bibitem{frankle2018lottery}
Jonathan Frankle and Michael Carbin,
\newblock ``The lottery ticket hypothesis: Finding sparse, trainable neural
  networks,''
\newblock in {\em ICLR}, 2019.

\bibitem{povey2018time}
Daniel Povey, Hossein Hadian, Pegah Ghahremani, Ke~Li, et~al.,
\newblock ``A time-restricted self-attention layer for asr,''
\newblock in {\em ICASSP}. IEEE, 2018, pp. 5874--5878.

\bibitem{zaheer2020big}
Manzil Zaheer, Guru Guruganesh, Kumar~Avinava Dubey, Joshua Ainslie, et~al.,
\newblock ``Big bird: Transformers for longer sequences,''
\newblock {\em Advances in Neural Information Processing Systems}, vol. 33, pp.
  17283--17297, 2020.

\bibitem{tay2020efficient}
Yi~Tay, Mostafa Dehghani, Dara Bahri, and Donald Metzler,
\newblock ``Efficient transformers: A survey,''
\newblock {\em ACM Computing Surveys (CSUR)}, 2020.

\bibitem{lee2022fithubert}
Yeonghyeon Lee, Kangwook Jang, Jahyun Goo, Youngmoon Jung, et~al.,
\newblock ``{FitHuBERT}: Going thinner and deeper for knowledge distillation of
  speech self-supervised models,''
\newblock in {\em Interspeech}, 2022, pp. 3588--3592.

\bibitem{wu2022performance}
Felix Wu, Kwangyoun Kim, Jing Pan, Kyu~J Han, et~al.,
\newblock ``Performance-efficiency trade-offs in unsupervised pre-training for
  speech recognition,''
\newblock in {\em ICASSP}. IEEE, 2022, pp. 7667--7671.

\bibitem{ours}
Yen Meng, Hsuan-Jui Chen, Jiatong Shi, Shinji Watanabe, et~al.,
\newblock ``On compressing sequences for self-supervised speech models,''
\newblock in {\em SLT}, 2022.

\bibitem{gao2022match}
Yan Gao, Javier Fernandez-Marques, Titouan Parcollet, Pedro~PB de~Gusmao,
  et~al.,
\newblock ``Match to win: Analysing sequences lengths for efficient
  self-supervised learning in speech and audio,''
\newblock in {\em SLT}, 2022.

\bibitem{vanhoucke2013multiframe}
Vincent Vanhoucke, Matthieu Devin, and Georg Heigold,
\newblock ``Multiframe deep neural networks for acoustic modeling,''
\newblock in {\em ICASSP}. IEEE, 2013, pp. 7582--7585.

\bibitem{miao2015eesen}
Yajie Miao, Mohammad Gowayyed, and Florian Metze,
\newblock ``{EESEN}: End-to-end speech recognition using deep {RNN} models and
  {WFST}-based decoding,''
\newblock in {\em ASRU}. IEEE, 2015, pp. 167--174.

\bibitem{chan2016listen}
William Chan, Navdeep Jaitly, Quoc Le, and Oriol Vinyals,
\newblock ``Listen, attend and spell: A neural network for large vocabulary
  conversational speech recognition,''
\newblock in {\em ICASSP}. IEEE, 2016, pp. 4960--4964.

\bibitem{vyas2022demand}
Apoorv Vyas, Wei-Ning Hsu, Michael Auli, and Alexei Baevski,
\newblock ``{On-demand compute reduction with stochastic wav2vec 2.0},''
\newblock in {\em Interspeech}, 2022, pp. 3048--3052.

\bibitem{cai2019once}
Han Cai, Chuang Gan, Tianzhe Wang, Zhekai Zhang, et~al.,
\newblock ``Once-for-all: Train one network and specialize it for efficient
  deployment,''
\newblock in {\em ICLR}, 2020.

\bibitem{wang2022lighthubert}
Rui Wang, Qibing Bai, Junyi Ao, Long Zhou, et~al.,
\newblock ``{LightHuBERT}: Lightweight and configurable speech representation
  learning with once-for-all hidden-unit {BERT},''
\newblock in {\em Interspeech}, 2022, pp. 1686--1690.

\bibitem{chang2022distilhubert}
Heng-Jui Chang, Shu-wen Yang, and Hung-yi Lee,
\newblock ``{DistilHuBERT: Speech representation learning by layer-wise
  distillation of hidden-unit BERT},''
\newblock in {\em ICASSP}. IEEE, 2022, pp. 7087--7091.

\bibitem{tsai2022superb}
Hsiang-Sheng Tsai, Heng-Jui Chang, Wen-Chin Huang, Zili Huang, et~al.,
\newblock ``{SUPERB-SG}: Enhanced speech processing universal performance
  benchmark for semantic and generative capabilities,''
\newblock in {\em ACL}, 2021.

\bibitem{dong2020cif}
Linhao Dong and Bo~Xu,
\newblock ``{CIF}: Continuous integrate-and-fire for end-to-end speech
  recognition,''
\newblock in {\em ICASSP}. IEEE, 2020, pp. 6079--6083.

\bibitem{panayotov2015librispeech}
Vassil Panayotov, Guoguo Chen, Daniel Povey, and Sanjeev Khudanpur,
\newblock ``Librispeech: an asr corpus based on public domain audio books,''
\newblock in {\em ICASSP}. IEEE, 2015, pp. 5206--5210.

\bibitem{liu2022towards}
Alexander~H. Liu, Wei-Ning Hsu, Michael Auli, and Alexei Baevski,
\newblock ``Towards end-to-end unsupervised speech recognition,''
\newblock in {\em SLT}, 2022.

\end{thebibliography}

\end{document}